\documentclass[conference,9pt]{IEEEtran}
\IEEEoverridecommandlockouts

\usepackage{cite}
\usepackage{amsmath,amssymb,amsfonts}
\usepackage{algorithmic}
\usepackage{graphicx}
\usepackage{textcomp}
\usepackage{xcolor}
\usepackage{comment}
\usepackage[numbers]{natbib}

\include{latex/macros}

\usepackage{hyperref}
\usepackage{mathtools}
\usepackage{multirow}
\usepackage{array}
\usepackage{colortbl}
\usepackage{footnote}
\usepackage{tablefootnote}
\definecolor{mygray}{gray}{.9}
\usepackage{pifont}
\newcommand{\cmark}{\ding{51}}%
\newcommand{\xmark}{\ding{55}}%

\def\BibTeX{{\rm B\kern-.05em{\sc i\kern-.025em b}\kern-.08em
    T\kern-.1667em\lower.7ex\hbox{E}\kern-.125emX}}
\begin{document}

\title{PI-Whisper: Designing an Adaptive and Incremental Automatic Speech Recognition System for Edge Devices 
}


\author{\IEEEauthorblockN{Amir Nassereldine$*$}
\IEEEauthorblockA{\textit{Computer Science and Engineering} \\
\textit{University at Buffalo}\\
Buffalo, USA \\
amirnass@buffalo.edu}
\and
\IEEEauthorblockN{Dancheng Liu$*$}
\IEEEauthorblockA{\textit{Computer Science and Engineering} \\
\textit{University at Buffalo}\\
Buffalo, USA \\
dliu37@buffalo.edu}
\and
\IEEEauthorblockN{Chenhui Xu}
\IEEEauthorblockA{\textit{Computer Science and Engineering} \\
\textit{University at Buffalo}\\
Buffalo, USA \\
cxu26@buffalo.edu}
\and
\IEEEauthorblockN{Ruiyang Qin}
\IEEEauthorblockA{\textit{Computer Science and Engineerning} \\
\textit{Unviersity of Notre Dame}\\
South Bend, USA \\
rqin@nd.edu}
\and
\IEEEauthorblockN{Yiyu Shi}
\IEEEauthorblockA{\textit{Computer Science and Engineerning} \\
\textit{Unviersity of Notre Dame}\\
South Bend, USA \\
yshi4@nd.edu}
\and
\IEEEauthorblockN{Jinjun Xiong}
\IEEEauthorblockA{\textit{Computer Science and Engineering} \\
\textit{University at Buffalo}\\
Buffalo, USA \\
jinjun@buffalo.edu}
}

\maketitle

\begin{abstract}

Edge-based automatic speech recognition (ASR) technologies are increasingly prevalent in the development of intelligent and personalized assistants. However, resource-constrained ASR models face significant challenges in adaptivity, incrementality, and inclusivity when faced with a diverse population. To tackle those challenges, we propose PI-Whisper, a novel ASR system that adaptively enhances recognition capabilities by identifying speakers' characteristics in real-time. In this work, we show how the design of PI-Whisper allows for incremental adaptation of new characteristics without the need for repetitive retraining, enhances recognition capabilities, and improves equity and fairness across diverse speaker groups.
PI-Whisper demonstrates these advantages by achieving state-of-the-art accuracy, reducing the word error rate (WER) by up to 13.7\% relative to baselines while scaling linearly to computing resources.
\end{abstract}


\section{Introduction}

With the increasing popularity of AI assistants, automatic speech recognition (ASR) is gaining growing attention from the research community~\cite{radford2022robust,bain23_interspeech}, especially its deployment onto edge devices when privacy is of great concern~\cite{liu2022private, qin2024empirical, qin2024enabling, qin2024robust, qin2024nvcim}. The recent transformer-based ASR models, such as Wav2Vec2~\cite{w2v2}, Conformer~\cite{gulati20_interspeech}, and Whisper~\cite{radford2022robust} have achieved great success.
However, the sizes of these cloud-based ASR models are significantly large and consist of many parameters, which poses practical challenges in computing power during fine-tuning and memory constraints during inference. These resource requirements are detrimental to the ASR models' deployment onto resource-constrained edge devices. To address these issues, people turn to the smaller versions of the ASR models, such as Whisper-tiny~\cite{radford2022robust}. 
But compared to the large models, these smaller models are less expressive and often struggle to provide accurate results
for speakers of diverse backgrounds.

In order for an intelligent assistant to truly deliver a personalized experience through ASR, three more important considerations must be addressed. \textbf{(1) The adaptivity of ASR to different speakers' characteristics as different users may have different ways of speaking the same languages.} For example, the characteristics of English speakers can be classified into various categories, including accents influenced by cultural or regional factors (such as African American, British, or Australian accents), age-related speech differences (such as the speech patterns of pre-schoolers compared to adults), and gender-specific speech differences (such as male versus female voices). Ideally, ASR should be adaptive to different speakers' characteristics to achieve higher accuracy, and this is especially true for edge-based ASR whose model sizes are constrained by the limited edge resources, as their expressivity limits their generalization capability. \textbf{(2) The incrementality of the ASR model to accommodate the ever-evolving interactions among different speakers.} For example, an intelligent assistant with an ASR system trained to understand general American English accents may later encounter situations where it needs to interact with speakers of various accents (e.g., Australian English). However, it is still unclear how to efficiently add new capabilities incrementally to an existing deployed ASR so that it can not only support existing speakers but also
support the new group of speakers without retraining the model. 
When the downstream task's distribution shifts, retraining ASR models from scratch is rarely feasible due to the quadratic cost with respect to the total training data size.
And \textbf{(3) the inclusivity of ASR to provide equal and fair experience for speakers with different characteristics.} In other words, how can we ensure the ASR model does not favor certain groups over others?

\begin{figure*}[h]
    \centering
    \includegraphics[width=1.0\textwidth]{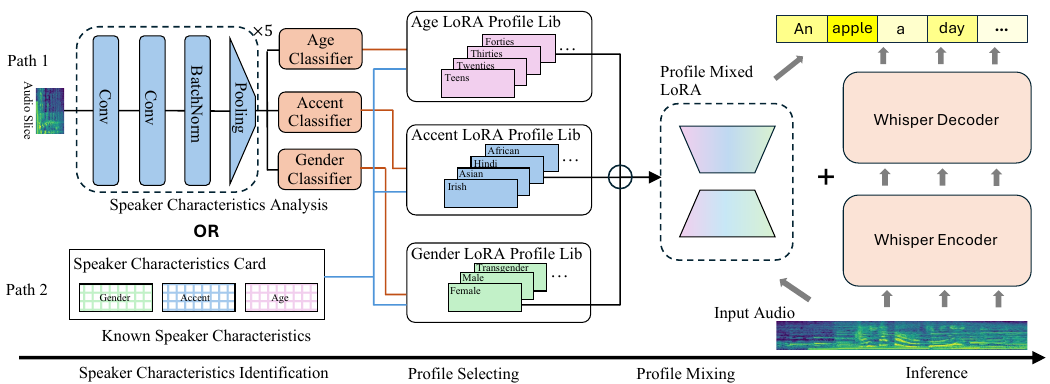}
    \caption{PI-Whisper leverages multiple LoRA profile libraries and dynamic LoRA merging of LoRA profiles to adjust ASR towards the speaker's characteristics. When the speakers' characteristics are not known, PI-Whisper employs a multi-head classifier to infer the characteristics from audio samples.}
    \label{fig:arch}
\end{figure*}

In this work, we propose a novel ASR system design, \textit{PI-Whisper}, to tackle these three challenges. As shown in Fig. \ref{fig:arch}, our system augments existing ASR frameworks with Low-Rank Adaptation (LoRA) profile libraries and an optional classifier. 
The LoRA profile library is a set of LoRA profiles organized by grouped speaker characteristics. During fine-tuning toward downstream tasks, these LoRA profiles are trained separately from ASR, and they are added and updated incrementally, providing a unique opportunity for the system to adaptively customize its capabilities based on speakers' characteristics. During inference, the dynamic merging of LoRA profiles before loading them onto the base ASR model can provide a fine-grained ASR service based on the speaker's unique characteristics. When the speaker's characteristics are unknown during inference, a classifier will provide an automated identification of speakers' characteristics so that PI-Whisper as a system can optimally choose the proper set of LoRA profiles for ASR customization. 

\noindent\textbf{Contributions} We make the following contributions:
\begin{itemize}
    \item We identify and address three key challenges in today's edge ASR services: adaptivity, incrementality, and inclusivity.
    \item We design PI-Whisper, a novel edge ASR system that augments pre-trained ASR models with characteristic classifiers and LoRA profiles. PI-Whisper serves as a robust solution that comprehensively targets the three aforementioned challenges.
    \item We show that human voices can be categorized into finite characteristic libraries, and PI-Whisper is able to use those characteristic libraries to provide accurate ASR services that generalize beyond training datasets.
\end{itemize}
Through extensive experiments, we show that PI-Whisper can not only achieve state-of-the-art ASR accuracy compared to existing works but also induce a minimal memory overhead for inference. PI-Whisper reduces the word error rate (WER) by up to relatively 13.7\% on the test dataset, with only 3.4\% inference time overhead.


\section{Background and Related Works}
\label{sec_background}

\subsection{ASR: Automatic Speech Recognition}

ASR is a technology that 
translates human speeches into a textual form for further downstream processing.
Over the years, significant advancements have been made in this field, driven by both academia and industry. 
Compared to early statistical models such as Kaldi \citep{Povey2011TheKS}, 
DNN-based ASR models, such as Recurrent Neural Networks (RNNs) \citep{Liu2021improving}, Long Short-Term Memory Networks (LSTMs) \citep{zeyer2017lstm}, and more recently, transformer-based models, have demonstrated remarkable improvements in recognition accuracy  \citep{gulati20_interspeech,w2v2,radford2022robust,bain23_interspeech}. Through large-scale pre-training with weak supervision, state-of-the-arts like Whisper \citep{radford2022robust} achieve robust and superior accuracy on most ASR tasks.


Nonetheless, one of the foremost practical challenges in ASR today is the balance between model accuracy and deployment feasibility. Arguably, transformer-based models all follow the scaling law \citep{kaplan2020scaling}, which states that the performance of a model is positively correlated to its model size. This phenomenon also holds true for ASR models, i.e., the larger the model is, the higher the accuracy is. This is due to the model's capacity to capture complex linguistic patterns and variations. Previous works have also observed that larger models of the Whisper family \citep{radford2022robust} on two benchmarks \citep{librispeech, Heikinheimo_2023} achieve higher quality transcriptions with lower word error rates (WER).

\subsection{ASR with Diverse Speech Characteristics}

Research has also focused on improving the robustness of ASR, which accepts additional speakers' characteristics as part of the input, typically the accent characteristics. On the one hand, there is research related to classifying speech into different accents, from the early works such as \cite{accent2006} to machine learning approaches such as \cite{word2016}, and to the recent deep learning approaches such as \cite{dl2022} and \cite{song2023mpsadensenet}. Overall, the advances in those models and algorithms have shown that accent classification is an achievable task with high accuracy. However, how those classification results could better guide the ASR model on the actual transcription task remains unclear. On the other hand, 
some previous works try to add accent information to the training stage of the model, but those works are less practical when the target population is not known \textit{a priori}
\citep{winata2020learning, zhou2023icassp}. 
When those models face a new accent, they need to be retrained for best quality, which involves the utilization of all previously known accents' data. \cite{prabhu2023accented} tries to address this issue of expensive retraining by developing an accent-robust model without using new accent data for training. While their approach provides some benefits when the new accents indeed have no training data, the improvement is very marginal (relatively 5\% compared to the previous state-of-the-art model \cite{das2021best}, as reported in the paper). Also, such an assumption of the lack of data is often not valid, and their model still needs to be retrained from scratch if they were to adapt to the new accents. 

\begin{table*}[ht]
\centering
   \tabcolsep 28pt
   \caption{Representative ASR works suffer from the inability to either target speakers from diverse backgrounds or to add more characteristics of the speakers (unless retraining with the whole dataset each time). Compared to them, PI-Whisper can seamlessly integrate any number of characteristics with incremental learning ability.}
\begin{tabular}{|c|ccc|}

\hline
Framework          & Deployment & Speaker Diversity  & Incremental Learning \\
\hline
Whisper large\footnote[1]{} \cite{radford2022robust}      & Cloud Only                         & \xmark                                                                           & --             \\
Whisper tiny  \cite{radford2022robust}     & Edge                           & \xmark                                                                          &    --          \\
MAML \cite{winata2020learning} & Edge                           & Accent                                                                       &  \xmark       \\
Accented Conformer \cite{prabhu2023accented} & Edge                           & Accent                                                                       & \xmark       \\
SAML \cite{zhao2024samlspeakeradaptivemixture} & Edge                           & Latent                                                                       & \xmark       \\
 \rowcolor{mygray}  PI-Whisper (ours)         & Edge                  & Accent+Gender+$\cdots$                                                                        & \cmark         \\
\hline
\end{tabular}
\label{tab:p-whisper-comparison}
\end{table*}

Albeit the efforts towards diverse accents, to the best of our knowledge, no existing edge-based ASR work can meet all the three outlined needs in the introduction yet, as summarized in Table \ref{tab:p-whisper-comparison}. Compared to existing works, PI-Whisper has two advantages. First, it is based on a linear-time non-intrusive approach. That is, the total training time scales linearly with the training data (as opposed to the quadratic total time for other approaches that require retraining), and fine-tuning with the proposed system does not need to change the original model's weights. Not changing the original model is especially important in the ASR field, as fine-tuning on one dataset will cause the model to lose generalizability and often perform worse on other unseen datasets. As a result, the pre-trained model should always be kept as the safe option. Second, it seamlessly integrates the potential multiple characteristics of the speaker. When the speaker has more information available besides his accent, our system utilizes every available characteristic (the profiles), whereas existing works fail to integrate multiple characteristics. In Section \ref{sec:experiments}, we will also show that incorporating additional characteristics yields a noticeable improvement in accuracy.

\subsection{LoRA Composition and Merging}
The origin of the LoRA merging technique is rooted in the area of diffusion models \cite{ho2020denoising}. The composition (or sometimes also called merging) of multiple LoRA profiles is an emerging technique that is widely used in image generation tasks. 
\cite{zhong2024multilora} has shown that LoRA composition is an excellent tool that retains all 
features of the target population with multiple attributes. However, up until now, such an approach was only proven to work with image generation tasks and very recently LLMs \citep{huang2024lorahub}. To the best of our knowledge, no previous work has explored the effectiveness of LoRA composition in ASR, which is more of a classification than a generation task. We are the first to show that such a non-intrusive approach also works for ASR.

\section{Problem Formulation}
\label{sec_prob}

This work focuses on developing an ASR inference system toward a non-static downstream distribution on the edge. In this section, we formally define the problem in mathematical notations.

First, we consider a system $M$ that contains some pre-trained ASR model on some generic distribution $G$ consisting of many audio and transcription pairs. Formally, $G$ can be denoted as $G=\{(a_i,t_i)\}_{i=1}^{N_G}$, where $a_i$ is the audio sample and $t_i$ is its corresponding transcription. The size of the pre-training dataset is $N_G$.

Since the key to improving downstream ASR tasks rests on the characteristics of the speaker, and since it is noted that a speaker may have multiple characteristics, we formally define characteristics as follows. For example, an Irish male speaker would possess both gender (male) and accent (Irish) characteristics. 
Formally, $\mathbf{C}$ is defined to be the set of speaker characteristics groups, such as $\mathbf{C}$ = \{Accent, Gender\}. It can be formalized further as
 $\mathbf{C} = \{\mathbf{C}^k\}_{k=1}^{|\mathbf{C}|}$, where
$|\mathbf{C}|$ gives the size of set  $\mathbf{C}$ and
$\mathbf{C}^k$ represents the $k$-th group of
speaker characteristics, such as Accent or Gender.
For each group of characteristics, it would have multiple possibilities, so $\mathbf{C}^k = \{c^{k,j}\}_{j=1}^{|\mathbf{C}^k|}$. 
For example, if we denote $\mathbf{C}^1$ as the Accent group, for some dataset.
it could have three accents in that group, i.e., $\mathbf{C}^1$ = \{Irish, American, Canadian\}. In other words,  $c^{1,1}$ would denote the Irish accent.

Based on the definitions above, we construct the definition of the fine-tuning dataset $D$ for the downstream task: 
$$D=\{d_i  \}_{i=1}^{N_D},$$

where $d_i=(a_i,t_i,\{c_i^{k,j}\})$ is a sample in the dataset with an audio $a_i$, its transcription $t_i$, and the speaker's characteristics as a set collection in $\{c_i^{k,j}\}$. In $\{c_i^{k,j}\}$ where $k$ indexes into the type of characteristic, and $j$ points to the exact characteristic. The $N_D$ = $|D|$ is the measure of the dataset.

When the fine-tuning dataset is a non-static dataset that keeps expanding with new data points, such non-static property can be represented by the time-varying nature of the set size of $N_D(T)$, moreover, $N_D(T)$ is a monotonically increasing function of time $T$, i.e., the length of the fine-tuning non-static dataset $D$ has the following property: $N_D(T)\leq N_D(T') \iff T \leq T'$.

Due to the dataset expansion, $\mathbf{C}$ and each $\mathbf{C}^k$ may also expand. Referring back to the example, after the dataset expands for some time, it is possible to have:
\begin{equation}
\begin{cases}
    \mathbf{C}=\text{\{Accent, Gender, \textit{Age}\}} \\
    \mathbf{C}^1=\text{\{Irish, American, Canadian, \textit{Australian}\}}
    \end{cases}
\label{eq.1}
\end{equation}

\footnotetext[1]{Whipser-large is diverse to speaker characteristics because of its strong generalization towards all speech, but it does not have additional features that explicitly support speaker characteristics that other works do.}

To maximize the accuracy of $M$ on the downstream task, the fine-tuning process uses whatever dataset at hand ($D$ with $T$=current) to fine-tune $M$. When the dataset expands and $ M$ needs to be adjusted again, the expanded dataset with a new $T'$ will be used.

During the inference stage, this work considers two different settings. First, similar to other models, $M$ only gets the audio $a_i$ without the speaker's characteristics, which we will refer to as ``Inferred Characteristics'' below. Under this setting, our objective will be $\min \textit{WER}[M(a_i),t_i]$. For the proposed PI-Whisper, it involves both inferring the speaker's characteristics and transcribing. In addition to this setting, we consider the case where the speaker's identity is known a priori, which we will refer to as ``Known Characteristics" below. We argue that this setting is also achievable under certain scenarios, particularly in most private settings where the ASR model is more of an assistant tool over a service. Under this setting, the objective function takes the speaker's characteristics as input, and it will be $\min \textit{WER}[M(a_i,\{c_i^{k,j}\}),t_i]$.

\vspace{3mm}
\section{PI-Whisper Design}
\label{sec_design}


In this section, we introduce \textbf{PI-Whisper}, which addresses the aforementioned limitations. By dissecting the fine-tuning from the model and the characteristics from the input, our proposed framework achieves linear scalability with training data and seamlessly supports any number of speaker characteristics. In particular, previous works need to use the entire $D$ at each update to $M$, but PI-Whisper only uses the new data after the last update. We will first introduce the overall architecture design, then dive into the details of individual components of the system during training, and finally describe the inference pipeline.

\subsection{PI-Whisper in a Nutshell}

As shown in Fig.~\ref{fig:arch}, the proposed system consists of three extra components added to the Whisper model. The first component
is used for speaker characteristic identification.
It is a CNN-like structure that consists of characteristic encoding blocks and many classifier heads. The speaker characteristic identification component takes a small sample of the audio input and identifies the speaker's characteristics such as age, accent, and gender. Based on the classified characteristics from the first component, the second PI-Whisper component will dynamically identify and retrieve the corresponding LoRA profile associated with each characteristic from the multiple LoRA profile libraries. Finally, in the last component, PI-Whisper concatenates the associated profiles and forms a distinctive merged LoRA profile that accurately represents the speech patterns of the target speaker. Benefiting from the merged LoRA profiles, the ASR model can achieve maximized transcription quality.

\subsection{The Training Pipeline}
\subsubsection{Data Processing}
Since PI-Whisper builds upon Whisper, we consider the same data processor as the one used by the original work. Through a discrete Fourier transform (DFT), the 30-second input audio is transformed into a log spectrogram with 80 features, in the shape of [80,3000]. We take a small subset of the transformed audio, e.g., the first 3 seconds of the input in the shape of [80,300] as the input to the encoder and classifier. Such a decision is made because the speaker's characteristics are very evident through even very few words \citep{word2016}, and our classifier is capable of classifying the characteristics with high accuracy, as shown in Section \ref{sec:experiments}.

\subsubsection{Characteristic Classifier}
As discussed in Section \ref{sec_background}, we find that there exist many methods to classify input speech samples into different characteristics. To showcase our system's capability, we choose to use a very simple VGG-style CNN network as the backbone architecture of the classifier \citep{simonyan2014very}. The proposed classifier has two parts, with a 10-layer CNN module as the feature encoder that is shared among different $\mathbf{C}^k$, and a classifier head for each of $\mathbf{C}^k$. The CNN module consists of 5 CNN blocks, each with two convolutional layers followed by a batch normalization and a pooling layer. The classifier head consists of three dense layers, mapping the latent features into $|\mathbf{C}^k|$. All layers in the classifier use ReLU as the activation function and cross-entropy as the loss function. Although different classifier designs would also work for our purposes, we chose such architecture as a proof of concept with minimal overhead. 

\subsubsection{LoRA Profile Library}
\label{sec:lora_choice}
Each type of characteristic, $C^k$, has a corresponding LoRA profile library that could be dynamically loaded into the memory and the framework. Within each library, each characteristic $c^{k,j}$ will have a corresponding LoRA profile. For ease of discussion, for the rest of the paper, we will use the notation of characteristics as the notation for LoRA profiles.


During training, when receiving a training sample $(a_i,t_i,\{c^{k,j}\})$, PI-Whisper will use the $(a_i,t_i)$ pair as the training data for each of $c^{k,j}_i \in \{c^{k,j}\}$. That is, suppose a training sample is from a speaker that has the following characteristics: \{Age: Teens, Accent: Irish, Gender: Female\}, then her audio and transcription pair will be used to train the Teens LoRA profile in the Age library, the Irish profile in the Accent library, and the Female profile in the Gender library.

\subsection{The Inference Pipeline}
During inference, the pipeline is slightly different depending on the problem setting for accessing the speaker's characteristics. Specifically, since the Inferred Characteristics setting does not have the speaker's characteristics, we need to rely on the classifier to infer the profiles that we need to pick from the libraries. Meanwhile, in the other setting, the ground truth of the speaker's characteristics is known beforehand, which means that we could directly use the ground truth as the label.

Upon knowing the characteristics, the next step is to choose the appropriate LoRA profiles from the libraries. Self-evidently, we choose the corresponding LoRA profiles based on the characteristics in the same way as defined by the training stage in Section \ref{sec:lora_choice}.


Afterward, we would have one profile from each of the $|\mathbf{C}|$ libraries, and we need to merge (or composite) them into one unified LoRA profile that could be used for the downstream task. 
During experiments, we have found that linearly summing up the weights according to Equation \ref{eq:linear} performs slightly worse than concatenating the LoRA profiles according to Equation \ref{eq:concat}. In the two equations, $W'(x)$ is the output from a layer with LoRA adaptation, $W(x)$ is the output from a layer calculated with weights from the pre-training, $w_j$ is the hyperparameter that controls the strength of the LoRA profile, where we follow the convention and set to be $\frac{1}{N}$, $\bigoplus$ is the concatenation, and $B_jA_j$ is the LoRA decomposition. 

\begin{equation}
    W'(x) = W(x)+[\sum_j^N w_j B_j A_j](x)
    \label{eq:linear}
\end{equation}
\begin{equation}
    W'(x) = W(x) + [ \bigoplus_{k=1}^{|\mathbf{C}|} w_k B_k \cdot\bigoplus_{k=1}^{|\mathbf{C}|} w_k A_k](x)
    \label{eq:concat}
\end{equation}

\subsection{Overhead Parameter Calculation}
To calculate the highest possible overhead $P_{\text{total}}$ of the PI-Whipser components, Equation \ref{eq: param} can be used as an estimate, where
 $P_{\text{enc}}$\footnote{It would be $P_{\text{enc}}$*$|\mathbf{C}|$ if each encoder is separated. In this paper, we consider the encoder to be shared across all classifier heads.} is the size of the CNN encoder, $P_{\text{h}}$ is the size of the classification head, $|\mathbf{C}|$ is the number of LoRA profile libraries, $P_{\text{pro}}$ is the size of each LoRA profile, and the summation counts all profiles in all libraries. 

\begin{equation}
\begin{split}
    P_{\text{total}} = P_{\text{enc}}+P_{\text{h}}*K+
    P_{\text{pro}}* \sum_{k=1}^{|\mathbf{C}|} |C^k|     
\end{split}
    \label{eq: param}
\end{equation}

This equation captures both settings regardless of whether the characteristics of the individual are known. In the Known Characteristics setting, there is no need for the classifier and hence the $P_{\text{enc}}$ and $P_{\text{h}}$ can be set to zero.
We will elaborate in Section \ref{sec:experiments} on the concrete parameter overhead and inference delay of the proposed system.

\section{Experiments and Results}
\label{sec:experiments}



\begin{table}[t]
\centering
\caption{Characteristics of L2-Arctic and CommonVoice Datasets}
\tabcolsep 9pt
\begin{tabular}{|c|c|c|c|}

\hline
Dataset & Gender & Accent & Age \\ \hline
L2-Arctic \citep{zhao18b_interspeech} & 
\begin{tabular}[c]{@{}c@{}}Male\\ Female\end{tabular} & 
\begin{tabular}[c]{@{}c@{}}Arabic (AR) \\ Mandarin (Ma) \\ Hindi (Hi) \\ Korean (Ko) \\ Spanish (Sp) \\ Vietnamese (Vi)\end{tabular} & 
-- \\ \hline
CommonVoice  \citep{ardila-etal-2020-common} & 
\begin{tabular}[c]{@{}c@{}}Male\\ Female\end{tabular} & 
\begin{tabular}[c]{@{}c@{}}Australian \\ English \\ Canadian \\ Scottish \\ United States\end{tabular} & 
\begin{tabular}[c]{@{}c@{}}Teens \\ Twenties \\ Thirties \\ ... \\ Nineties\end{tabular} \\ \hline
\end{tabular}
\label{tab:char}
\end{table}

\subsection{Experiment Setup}

\textbf{Datasets:} For our experiments, we use two datasets: the L2-Arctic dataset (v5 release) \citep{zhao18b_interspeech} and a subset of the CommonVoice dataset (v17) \citep{ardila-etal-2020-common}, specifically the same subset utilized by \cite{prabhu2023accented}
. 
The two datasets include audio samples and their corresponding speakers' characteristics such as gender and accents. 
Table \ref{tab:char} summarizes
the groups of speaker characteristics and 
the unique profiles under each characteristic group for both datasets.   
For any train-val-test split, we consider a 50:20:30 split. 

\textbf{Models:} 
We use the SGD optimizer to train the speaker characteristics classifier for 10 epochs with a learning rate of $1 \times 10^{-4}$ and a batch size of 256. For the base ASR model, we employ Whisper-tiny \citep{radford2022robust}, a lightweight model with 37.7 million parameters, suitable for most edge devices. During fine-tuning with LoRA, we perform a hyperparameter grid search over learning rates \{1e-3,1e-4,5e-4, 1e-5,5e-5\}, and fine-tune the model for 3 epochs using the Huggingface Trainer \citep{Wolf_Transformers_State-of-the-Art_Natural_2020}, keeping all other hyperparameters at their default values.

The LoRA profiles are fine-tuned over the query and key projections of both the encoder and decoder in the Whisper model, selecting the best configuration based on validation dataset performance. Word Error Rate (WER) is used as the evaluation metric for transcription accuracy. Training and fine-tuning of the classifiers and the LoRA Libraries are conducted on an Nvidia A6000 GPU server, while inference times are evaluated on two edge devices: Raspberry Pi 5 and Jetson Orin Nano.


\begin{figure}[t]
    \centering
    \includegraphics[width=0.49\textwidth]{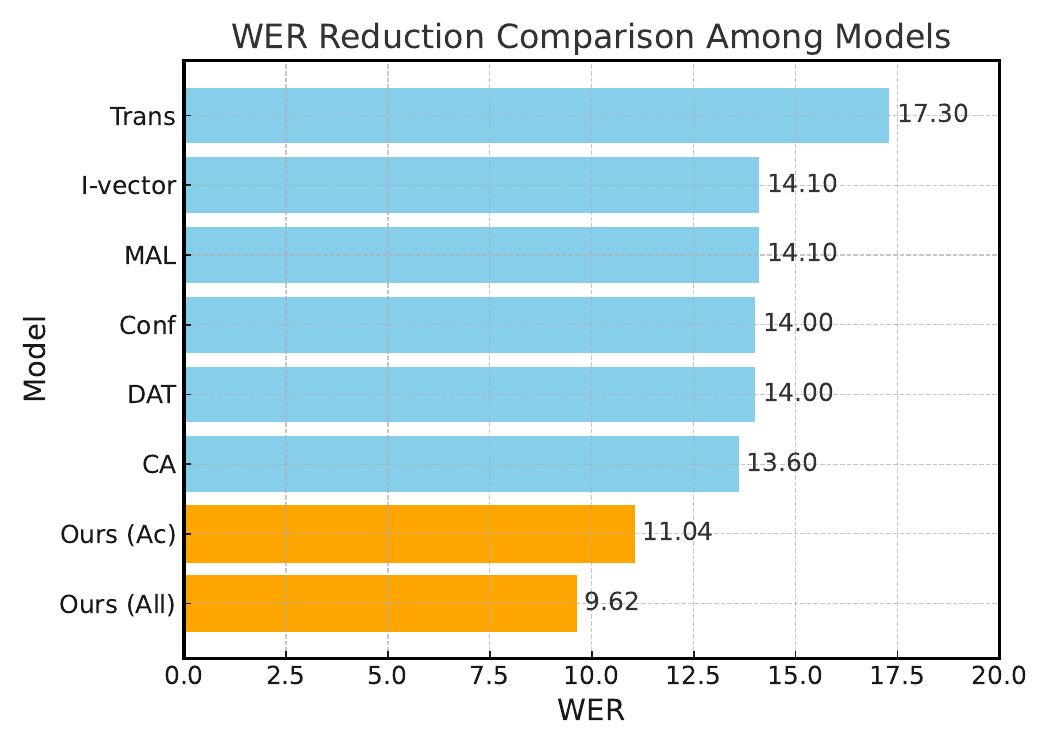}
    \caption{Comparison between PI-Whisper (in orange) and other baseline models on the CommonVoice dataset. All baseline results (in blue) are obtained from \cite{prabhu2023accented}. Ours (Ac) is the PI-Whisper with the accent profile library only, while Ours (All) is the PI-Whisper with all three profile libraries, namely accent, gender, and age.}
    \label{fig:baseline_comp}

\end{figure}
\subsection{Comparative Results for PI-Whisper}
In this section, we compare PI-Whisper under Inferred Characteristics setting with other existing ASR models using WER as the metric. 
While the Whisper model family is currently the most robust model, other models, especially the Conformer model family as reported in \citep{gulati20_interspeech}, could be trained or fine-tuned towards downstream tasks and achieve better performance. Thus, we compare our work with CA \citep{prabhu2023accented}, the most recent Conformer variation available. Since we are not able to reproduce their code on our platform, we report the numbers directly from their paper as well as their baselines, i.e., Trans \citep{dong2018}, Conf  \citep{gulati20_interspeech}, I-vector  \citep{Chen2015ImprovingDN}, MAL  \citep{zhang2021e2ebased}, and DAT \citep{das2021best}. We summarize the comparison results on the CommonVoice subset in Figure \ref{fig:baseline_comp}. As we can see from the figure, PI-Whisper achieves new SOTA WER results, and the reduction is 3.98\% compared to the existing best results as represented by CA.

\subsection{Ablation Study}
\begin{table}[t]
\centering
   \tabcolsep 5pt
   \caption{PI-Whisper performance compared to the untuned model (Base Model) and one LoRA fine-tuned on the whole dataset (One for All).}
\begin{tabular}[width=0.45\textwidth]{|p{0.07\textwidth}|p{0.135\textwidth}|p{0.067\textwidth} p{0.085\textwidth}|}
\hline
                                   &       \centering WER      & \centering L2 Arctic                   & \centering\arraybackslash Common Voice \\ 
\cline{2-4}
\multicolumn{1}{|c|}{\multirow{2}{*}{\hfil Baseline}}           & \centering Base Model        &    \centering 24.51                   &     \centering\arraybackslash 13.03   \\
                                   & \centering One for All    &     \centering 20.23                   &      \centering\arraybackslash 10.27       \\ \hline
\multicolumn{1}{|c|}{\multirow{7}{*}{\shortstack{PI-Whisper\\(Known\\Characteristics)}} }         & \centering Accent (Ac)  &     \centering 17.67                   &     \centering\arraybackslash 10.26   \\
                                   & \centering Gender (Ge)  &     \centering 19.52                   &     \centering\arraybackslash 9.87    \\
                                   & \centering Age (Ag)     & \centering --                          &     \centering\arraybackslash  10.45   \\
                                   & \centering Ac + Ge      &     \centering \textbf{17.45}          &     \centering\arraybackslash 9.83    \\
                                   & \centering Ac + Ag      & \centering --                          &  \centering\arraybackslash     9.83     \\
                                   & \centering Ge + Ag      & \centering --                          &    \centering\arraybackslash    9.51     \\
                                   & \centering Ac + Ge + Ag & \centering --                          &   \centering\arraybackslash         \textbf{9.24} \\ \hline
\multicolumn{1}{|c|}{\multirow{7}{*}{\shortstack{PI-Whisper\\(Inferred\\Characteristics)} }}          & \centering Accent (Ac)  &     \centering 17.65                        &    \centering\arraybackslash     11.04          \\
                                   & \centering Gender (Ge)  &             \centering19.54                &      \centering\arraybackslash   9.92     \\
                                   & \centering Age (Ag)     & \centering --                          &       \centering\arraybackslash       10.73 \\
                                   & \centering Ac + Ge      &                \centering \textbf{17.57}         &  \centering\arraybackslash    9.65            \\
                                   & \centering Ac + Ag      & \centering --                          &      \centering\arraybackslash     9.88   \\
                                   & \centering Ge + Ag      & \centering --                          &      \centering\arraybackslash     9.63   \\
                                   & \centering Ac + Ge + Ag & \centering --                          &      \centering\arraybackslash   \textbf{9.62}     \\ \hline
\end{tabular}

\label{tab:p-whisper_with_no}
\end{table}

Next, we show that using multiple LoRA profiles indeed brings benefits to the ASR model's performance. The results are summarized in Table \ref{tab:p-whisper_with_no}, and it is easy to conclude that PI-Whisper is an effective ASR system optimized for handling diverse speaker characteristics.
\begin{table}[!h]
\centering
\tabcolsep 22pt
\caption{Accuracy of speaker characteristics classifiers.}
\begin{tabular}{|c|cc|}
\hline
Accuracy & L2-Arctic & CommonVoice \\
\hline
Accent   & 98\%                            & 85\%                                                                             \\
Gender   & 99\%                            & 98\%                                                                             \\
Age      & --                            & 79\%             \\
\hline
\end{tabular}

\label{tab:classifier}
\end{table}

First, while the Whisper-tiny model does gain substantial improvement after fine-tuning with a single LoRA profile (One for All), PI-Whisper can even further improve from that with more fine-grained LoRA profile libraries. For the L2-Arctic dataset, merging accent and gender profiles gives up to 2.78\% (13.7\% relative) improvement in WER compared to the model with a single LoRA, while the improvement for CommonVoice is 1.03\% (10.0\% relative).

Second, merging LoRA profiles is \textit{always} beneficial for ASR. The more profiles, the better the accuracy we can get with PI-Whisper.

Lastly, it is interesting to note that 
PI-Whisper achieves comparable WER results
for the known (given) speaker characteristics
and the inferred speaker characteristics (through our classifiers). We attribute
such comparable WER results to
the high accuracy of our trained speaker characteristics identification classifiers.
We show such results in
Table \ref{tab:classifier}. As we can see from the table, the classifier performs especially well for relatively easy speaker characteristics like gender. 


\subsection{Overhead Analysis}
\label{sec:overhead}
\begin{figure}[h]
    \centering
    \includegraphics[width=0.45\textwidth,height=5cm]{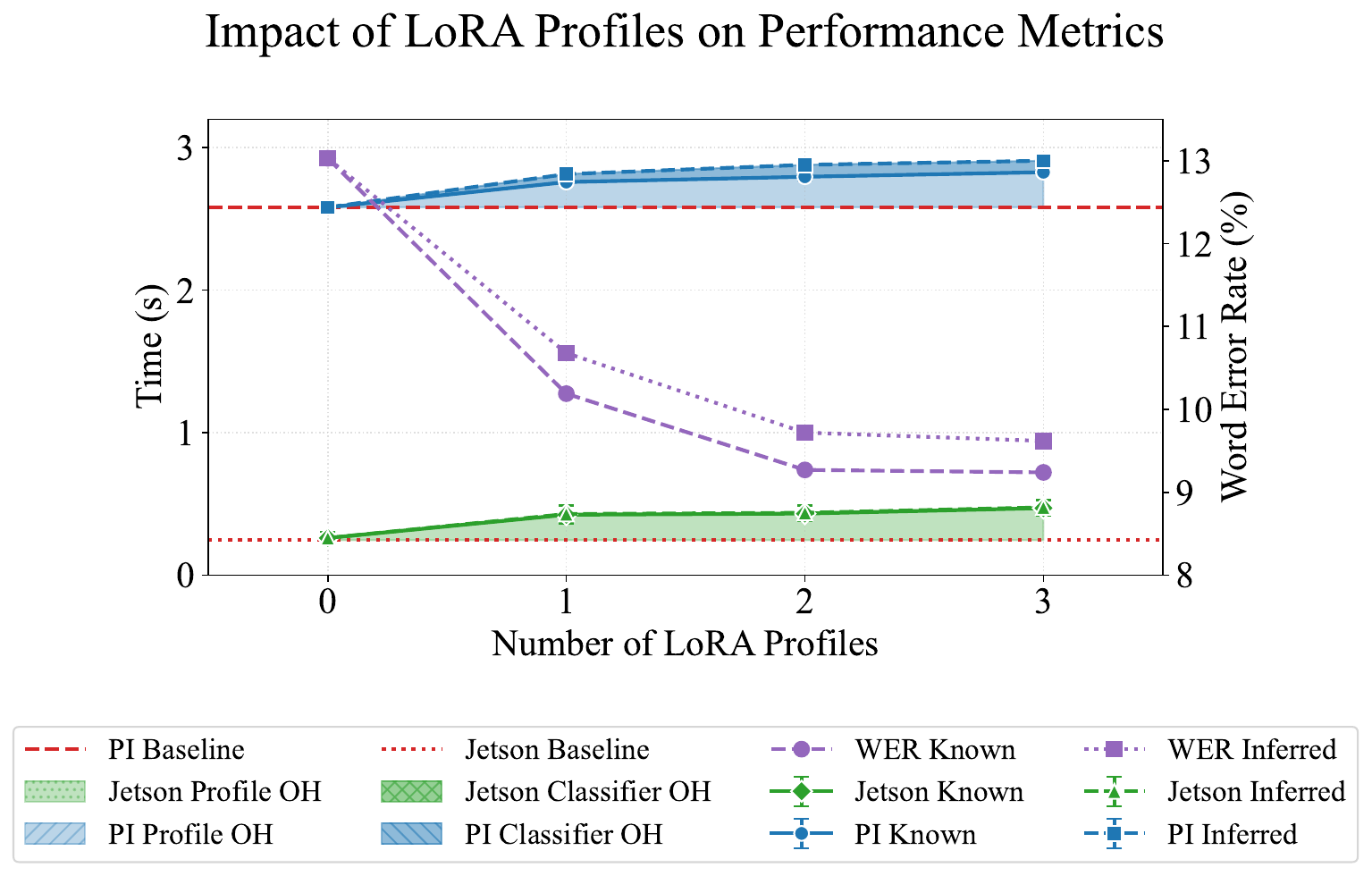}
    \caption{Impact of LoRA profiles on inference time and Word Error Rate (WER) for Raspberry Pi and Jetson devices, highlighting baseline performance, overhead contributions, and WER trends across Known and Inferred settings. }
    \label{fig:inference_speed}

\end{figure}

In this section, we will show the runtime performance of PI-Whisper compared to the baselines. Since we choose to use concatenation as the LoRA composition method, we report the usage over multiple LoRA profiles. Using Raspberry Pi 5 and Jetson Orin Nano, we run each setup on 100 samples. In Figure \ref{fig:inference_speed}, we show the tradeoff between inference time and WER on both hardware under both settings (known and Inferred). From the results, we draw some noteworthy observations:
\begin{itemize}

    \item The inference time scales linearly with the number of profiles for both settings on both devices. Adding one LoRA profile incurs \textbf{0.037} seconds of additional inference time on Raspberry Pi and \textbf{0.023} seconds on Jetson.

    \item The overhead introduced by using LoRA as the fine-tuning method, compared to no fine-tuning, is minimal and acceptable, approximately \textbf{0.18 seconds}.

    \item The classifier overhead is also within acceptable limits, adding approximately \textbf{0.056} seconds per sample.
\end{itemize}

\begin{figure}
    \centering
    \includegraphics[width=0.45\textwidth,height=5cm]{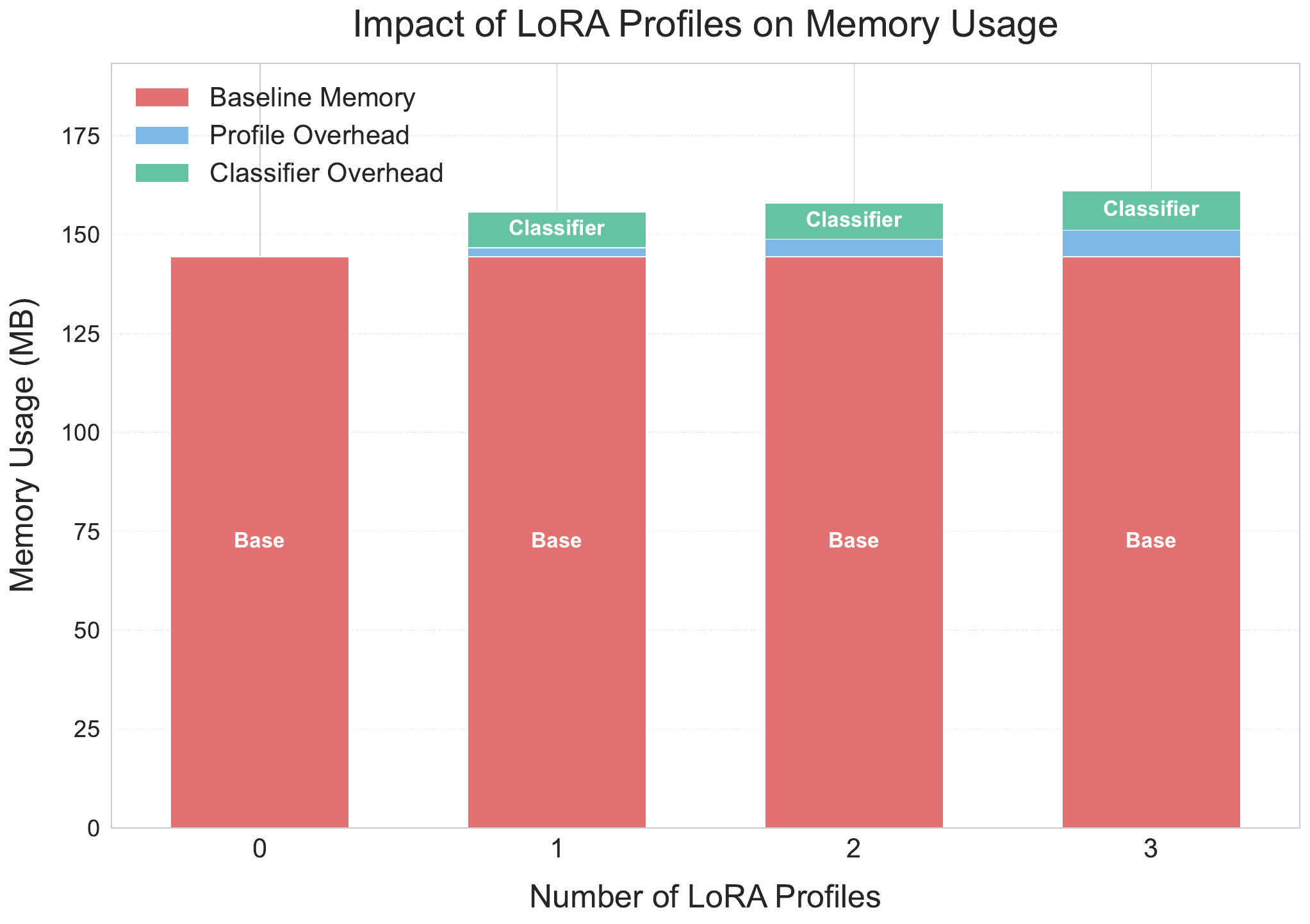}
  \caption{Impact of LoRA profiles on memory usage with dynamic profile loading, illustrating the breakdown of baseline memory, profile overhead, and classifier overhead as the number of profiles increases.}
\vspace{-3mm}
    \label{fig:memory}
\end{figure}

The marginal overhead in inference delay from LoRA and CNN classifiers is reasonable because the main bottleneck for inference is the attention layers in the transformer architecture \citep{ermis2022memory}. Based on the parameter sizes and the total number of LoRA profiles, the theoretical highest RAM overhead is around 21.8\% for L2-Arctic and 37.4\% for CommonVoice in the Inferred Characteristics setting, and 12.5\% for L2-Arctic and 25.0\% for CommonVoice in the Known Characteristics setting if all profiles and classifier were to be preloaded into memory. However, this overhead can be reduced by dynamically loading/unloading the required LoRA profiles during inference, as presented in Fig.  \ref{fig:memory}. 
Given that PI-Whisper is targeted towards edge devices, this setup provides an ideal setting for such devices where memory is scarce.
 Concretely, during the experiments, we observed memory overhead as shown in Fig. \ref{fig:memory}. The overhead of running PI-Whisper is around 1.55\% for a profile and 6\% for a classifier, which is expected from Equation \ref{eq: param}. 

\subsection{Zero Shot Learning}

In addition to our previous experiments, we performed zero-shot evaluations on the L2-Arctic dataset to study the efficacy of PI-Whisper in transferring knowledge. We train our LoRA libraries and classifiers on the CommonVoice dataset and then evaluate them on the L2-Arctic. The results in Table \ref{tab:zeroshot}  demonstrate that PI-Whisper significantly outperforms previous methods as well as the baseline approach, where a single LoRA is trained on CommonVoice (One for All). These findings indicate that PI-Whisper can effectively transfer knowledge across datasets, and the merging of different LoRA profiles enhances accuracy even further. We can also observe that traditional fine-tuning on one dataset loses generalization when it is zero-shot tested on another dataset (26.7\% vs 24.51\% for the Base Model). This is expected because fine-tuning with a single LoRA optimizes the model towards some specific features existing in the fine-tuning dataset, reducing the generalization of the model. Similar results have been reported by \cite{liu2024automatic}. However, PI-Whisper takes advantage of the generic characteristics across datasets, which allows a much more robust representation of the features. It can also be noted that the gender classifier which was trained on CommonVoice remains robust (81\% accuracy) on the L2-Arctic dataset.


\begin{table}[h]
\centering
\caption{Comparison of the Zero Shot WER\% of PI-Whisper with \cite{prabhu2023accented} on the L2-Arctic dataset.} 
\resizebox{\columnwidth}{!}{
\begin{tabular}{|c|c|cccccc|}
\hline
\multirow{2}{*}{Method} & \multirow{2}{*}{All} & \multicolumn{6}{c|}{Accents}                                                                                                           \\ \cline{3-8} 
                        &                      & \multicolumn{1}{c|}{Ar} & \multicolumn{1}{c|}{Hi} & \multicolumn{1}{c|}{Ko} & \multicolumn{1}{c|}{Ma} & \multicolumn{1}{c|}{Sp} & Vi   \\ \hline
CA \citep{prabhu2023accented}            & 32.6                 & 29.5                    & 30.4                    & 26.2                    & 37.1                    & 29.3                    & 42.8 \\
One for All             & 26.70                 &           23.00              &  14.59                       &    16.13                     &             25.71            &      27.14                   &    35.49  \\ \hline
PI-Whisper (Ge)         & 22.24                &                   21.08      &     14.16                    &    \textbf{16.10 }                    &                26.48        &      20.20             &    35.06  \\
PI-Whisper (Ag)         &   22.30                   &          20.84               &   14.88                      &    17.05                     &  26.21                       &    19.70                     &   34.71   \\
PI-Whisper (Ge + Ag)    & \textbf{21.67}       &       \textbf{20.27 }                 &     \textbf{  13.19  }                &     18.11                    & \textbf{ 25.06 }                      &     \textbf{19.03}                    & \textbf{33.9}      \\ \hline
\end{tabular}
}

\label{tab:zeroshot}

\end{table}

\subsection{Fairness Analysis}
Last but not least, since PI-Whisper can adapt to different speakers' characteristics, we expect it to perform equally well for different speaker groups without significantly biasing toward certain diversity groups. To validate such a hypothesis,
 we run a set of empirical experiments on the CommonVoice dataset and measure the effect of using multiple LoRA profiles. We adopt two fairness metrics, the statistical parity difference \textit{(SPD)} and disparate impact ratio \textit{(DIR)} \citep{fairness2023}, while considering the use of our WER metric as shown in Equation \ref{eq:spd} and \ref{eq:dir}. SPD is defined as the difference in WER between the best and the worst speaker groups; while
DIR is defined as the ratio between the best and worst speaker groups.

\begin{equation}
    \text{SPD} = \max |\text{WER}_i - \text{WER}_j| \quad \forall i,j \in C^k
    \label{eq:spd}
\end{equation}
\begin{equation}
    \text{DIR} = \max |\frac{\text{WER}_i}{\text{WER}_j}| \quad \forall i,j \in C^k
    \label{eq:dir}
\end{equation}

We report the results in Table \ref{tab:fairness_combined}. We first observe that, compared to the base model and the fine-tuned model with the traditional LoRA approach, our solution with a LoRA profile library for a specific speaker group characteristic can achieve higher fairness scores. The fairness scores get even better when more than one LoRA profile 
are considered, which is consistent with our expectations.

\begin{table}[h]
\centering
\tabcolsep 12pt

\caption{Fairness between using PI-Whisper LoRA profiles, using an untuned model (Base Model), and one LoRA fine-tuned on the whole dataset (One for All) on CommonVoice.} 
\begin{tabular}{|c|c|cc|}
\hline
Characteristic  & Model & SPD & DIR \\
\hline
\multirow{4}{*}{Accent} & Base Model & 0.5028 & 5.9696 \\
                        & One for All & 0.2843 & 4.4919 \\
                        & PI-Whisper (Ac) & 0.2620 & 4.0154 \\
                        & PI-Whisper (All) & \textbf{0.2065} & \textbf{3.6375} \\
\hline
\multirow{4}{*}{Gender} & Base Model & 0.0408 & 1.3647 \\
                        & One for All & 0.0245 & 1.2677 \\
                        & PI-Whisper (Ge) & 0.0254 & 1.2746 \\
                        & PI-Whisper (All) & \textbf{0.0222} & \textbf{1.2545} \\
\hline
\multirow{4}{*}{Age}    & Base Model & 0.0833 & 1.9292 \\
                        & One for All & 0.0588 & 1.7959 \\
                        & PI-Whisper (Ag) & 0.0578 & 1.6988 \\
                        & PI-Whisper (All) & \textbf{0.0442} & \textbf{1.6266} \\
\hline
\end{tabular}

\label{tab:fairness_combined}

\end{table}


\section{Conclusion}
\label{sec_conclusion}

In conclusion, we propose a novel ASR system, PI-Whisper. Compared to existing ASR systems, PI-Whisper has advantages in deployment feasibility, adherence to speaker diversity, and the ability to incrementally adjust to new speaker characteristics. By leveraging multiple LoRA profile libraries and dynamically selecting and merging the LoRA profiles based on the characteristics of the speaker, PI-Whisper provides better performance for downstream ASR tasks in terms of transcription accuracy and fairness. 

\renewcommand{\bibfont}{\small}
\bibliographystyle{ieeetran}
\bibliography{sample-base}

\end{document}